%% file: acl_latex.tex
\documentclass[11pt]{article}

\usepackage[preprint]{acl}

\usepackage{times}
\usepackage{latexsym}

\usepackage[T1]{fontenc}

\usepackage[utf8]{inputenc}

\usepackage{microtype}

\usepackage{inconsolata}

\usepackage{graphicx}

\usepackage{subcaption}
\usepackage{booktabs}
\usepackage{hyperref}
\usepackage[dvipsnames,table]{xcolor}
\usepackage{amsmath}
\usepackage{amssymb}
\usepackage{mathtools}
\usepackage{amsthm}
\usepackage[capitalize,noabbrev]{cleveref}
\usepackage{makecell}
\usepackage{arydshln}
\usepackage{multirow,multicol}

\newcommand{\modelname}{\textsc{HinT-SD}}
\newcommand{\abs}[1]{\left| #1 \right|}

\definecolor{myblue}{HTML}{E6F3FF}

\usepackage{tcolorbox}

\usepackage{listings}
\tcbuselibrary{listings}
\newtcblisting{promptbox}[1][]{
    coltitle=white,
    listing only,
    boxed title style={sharp corners, frame hidden},
    fonttitle=\ttfamily\centering\bfseries,
    title={#1},            
    listing options={
        language={},
        basicstyle=\small\ttfamily,
        breaklines=true,
        breakautoindent=false,
        breakindent=0pt,
        columns=fullflexible,
        showstringspaces=false,
        escapeinside={(*}{*)},
    },
    boxsep=2pt,
    top=2pt,
    bottom=2pt,
}

\tcbuselibrary{skins} 
\tcbuselibrary{breakable}
\definecolor{targettaskbg}{HTML}{FFF8E1}
\definecolor{targettaskframe}{HTML}{F9A825}
\definecolor{targettrajbg}{HTML}{E8F5E9}
\definecolor{targettrajframe}{HTML}{388E3C}
\definecolor{targetfeedbackbg}{HTML}{E3F2FD}
\definecolor{targetfeedbackframe}{HTML}{1565C0}
\definecolor{targetmetabg}{HTML}{F3E5F5}
\definecolor{targetmetaframe}{HTML}{7B1FA2}
\definecolor{targettoolcall}{HTML}{0D47A1}
\definecolor{targettoolresult}{HTML}{1B5E20}
\definecolor{targettag}{HTML}{BF360C}
\tcbset{
  targetmetastyle/.style={enhanced,colback=targetmetabg,colframe=targetmetaframe,left=4pt,right=4pt,top=3pt,bottom=3pt,boxrule=0.8pt,arc=2pt,fontupper=\small\bfseries},
  targettaskstyle/.style={enhanced,breakable,colback=targettaskbg,colframe=targettaskframe,fonttitle=\bfseries\small,left=4pt,right=4pt,top=3pt,bottom=3pt,boxrule=0.8pt,arc=2pt,fontupper=\small},
  targettrajstyle/.style={enhanced,breakable,colback=targettrajbg,colframe=targettrajframe,fonttitle=\bfseries\small,left=4pt,right=4pt,top=3pt,bottom=3pt,boxrule=0.8pt,arc=2pt,fontupper=\footnotesize\ttfamily},
  targetfeedbackstyle/.style={enhanced,breakable,colback=targetfeedbackbg,colframe=targetfeedbackframe,fonttitle=\bfseries\small,left=4pt,right=4pt,top=3pt,bottom=3pt,boxrule=0.8pt,arc=2pt,fontupper=\small}
}

\title{\modelname{}: Targeted Hindsight Self-Distillation for Long-Horizon Agents}

\author{
    Woongyeong Yeo$^{1*}$ \;
    Yumin Choi$^{1*}$ \;
    Taekyung Ki$^{1}$ \;
    Sung Ju Hwang$^{1,2}$ \\
    $^{1}$KAIST \;\; $^{2}$DeepAuto.ai \\
    \texttt{\{wgcyeo, yuminchoi, taekyung.ki, sungju.hwang\}@kaist.ac.kr} \\
    \url{https://github.com/wgcyeo/HINT-SD}
}

\begin{document}
\maketitle

{
    \renewcommand\thefootnote{*}
    \footnotetext{Equal contribution}
}

\input{sections/0_abstract}
\input{sections/1_introduction}
\input{sections/2_method}
\input{sections/3_experiment}

\input{sections/4_conclusion}

\input{sections/X_limitations}
\input{sections/X_ethical_considerations}
\input{sections/X_acknowledgements}

\bibliography{reference}

\input{sections/X_appendix}

\end{document}

%% file: sections/0_abstract.tex
\begin{abstract}
Training long-horizon LLM agents with reinforcement learning is challenging because sparse outcome rewards reveal whether a task succeeds, but not which intermediate actions caused the outcome or how they should be corrected. Recent methods alleviate this issue by generating rewards or textual hints from turn-level action-output signals, or by using feedback-conditioned self-distillation. However, generating feedback at every turn is inefficient when many intermediate turns are already successful or neutral, and applying feedback at a fixed or misaligned turn often fails to supervise the actions that contributed to the failure. To bridge this gap, we propose \modelname{}, a targeted self-distillation framework that uses full-trajectory hindsight to select failure-relevant actions and applies feedback-conditioned distillation only to targeted action spans. Experiments on BFCL v3 and AppWorld show that our method outperforms the dense per-turn feedback baseline by up to 13.60 percentage points on average while achieving a 2.26$\times$ reduction in time per training step, suggesting that selecting where to distill is key to effective and efficient long-horizon agent training.
\end{abstract}

%% file: sections/1_introduction.tex
\section{Introduction}

Large language model (LLM) agents are widely adopted to automate complex workflows through long-horizon interactions with tools, APIs, and web interfaces~\citep{yao2023react,zhou2024webarena,trivedi2024appworld,bfcl}. Reinforcement learning (RL) has become an increasingly common post-training paradigm for improving agents from verifiable task results. However, long-horizon agent tasks typically provide only sparse, binary rewards that indicate whether the task succeeded, offering limited guidance on which intermediate decisions contributed to success or failure or how agent behavior should be improved.

\input{figures/concept_figure}

Recent work has addressed this sparsity by providing denser learning signals. AgentEvolver~\citep{agentevolver} augments GRPO with LLM-based self-attribution, assigning process-level contribution rewards to intermediate actions. SDPO~\citep{hubotter2026sdpo} and RLTF~\citep{song2026rltf} use rich feedback as privileged training context, distilling feedback-conditioned teacher distributions from textual critiques, runtime errors, or failed tests into a feedback-free student. OpenClaw-RL~\citep{wang2026openclaw} extends this idea to agentic interactions by using next-state signals such as user replies, tool outputs, and environment transitions to generate turn-level rewards and textual hints for on-policy distillation.

While providing denser feedback, these approaches leave its localization unresolved, making corrective supervision highly ineffective for long-horizon agents.
For instance, although self-attribution can identify failure-causing actions, its signal remains a scalar reward, making the agent still rely on sparse successful rollouts to learn the correct alternative. Feedback-conditioned distillation provides token-level teacher supervision, but applying hindsight feedback before the first action or distilling the full trajectory can misalign the teacher and student. After the erroneous turn identified by feedback, the student's subsequent trajectory may already deviate from the trajectory supported by the feedback-conditioned teacher, making later token targets unreliable and dominated by accumulated mismatch rather than the intended local correction. OpenClaw-RL localizes feedback to each action, but it must evaluate every turn and remains tied to immediate action-output transitions, making delayed failures difficult to attribute. The remaining bottleneck is therefore not only how to obtain richer feedback, but also how to place it on the action spans where it is relevant.

We view this as a \emph{relevance-sparsity} problem: in a failed trajectory, only a small subset of actions may require correction, since most turns are either correct or consequences of earlier mistakes. Supervising all turns wastes training budget and can introduce noisy updates. Moreover, in long-horizon interactions, errors often become visible after their causal actions. Applying feedback at the observed failure can therefore supervise downstream consequences instead of the original mistakes. Effective hindsight learning should first identify where feedback is relevant before using it for policy updates.

We propose \modelname{}, a self-distillation framework that converts hindsight feedback into targeted token-level supervision. Given a failed trajectory, \modelname{} analyzes the full rollout to produce a sparse set of failure-relevant steps together with corrective feedback for each step. For each selected step, the same policy serves as a hindsight-conditioned teacher by observing the original prefix plus the generated feedback, while the student observes only the original prefix. \modelname{} then applies a distillation loss only to the selected action spans, encouraging the student to internalize corrective behavior.

Our contributions are as follows: (i) we identify relevance-sparsity as a key obstacle in long-horizon agent training and formulate hindsight distillation as a target-selection problem; (ii) we propose \modelname{}, a self-distillation framework for long-horizon agent training that distills a feedback-conditioned teacher only at selected failure-relevant actions; (iii) on BFCL v3 and AppWorld, \modelname{} outperforms the dense per-turn feedback baseline by up to 13.60 percentage points on average while reducing per-step training time by 2.26$\times$.

%% file: figures/concept_figure.tex
\begin{figure}[t!]
    \centering
    \includegraphics[width=\linewidth]{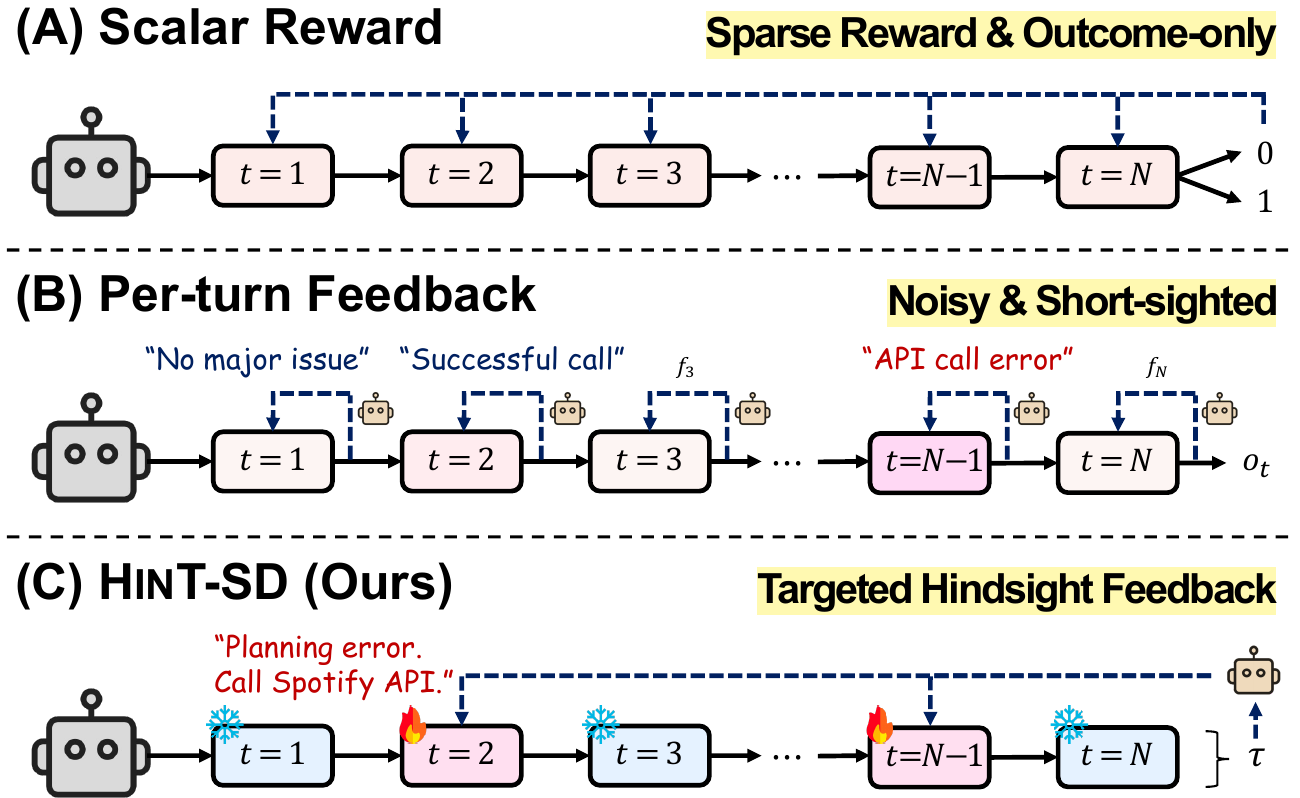}
    \vspace{-0.2in}
    \caption{Comparison of strategies for long-horizon agent training. \modelname{} uses full-trajectory hindsight to identify failure-relevant turns and applies feedback-conditioned distillation only to the corresponding action spans, avoiding sparse or mislocalized supervision.}
    \vspace{-0.2in}
    \label{fig:concept}
\end{figure}

%% file: sections/2_method.tex
\section{Method}

\paragraph{Problem Setup.}
We consider a multi-turn agent policy $\pi_\theta$ interacting with an environment over a trajectory
\begin{equation}
    \tau = (s_1, a_1, \cdots, s_T, a_T), \label{eq:trajectory}
\end{equation}
where $s_t$ denotes the environment state observed at step $t$, which may include tool outputs, error messages, and other interaction feedback. At each step, the agent samples an action $a_t \sim \pi_\theta (\, \cdot \mid h_t)$ conditioned on the interaction history $h_t=(s_1, a_1, \dots, s_t)$. Our goal is to improve task outcomes by applying supervision only to action spans responsible for failure while preserving useful behavior elsewhere.

\paragraph{Overview.}
To address the challenges of sparse trajectory-level rewards and the inefficiency of uniform process supervision, we propose \modelname{}, a targeted self-distillation framework guided by self-generated hindsight. Given a failed rollout, \modelname{} analyzes the complete trajectory in hindsight to isolate a set of failure-relevant actions, and then applies feedback-conditioned self-distillation only to the token spans of these selected actions.

\paragraph{Hindsight Feedback Generation.}
Identifying the true source of failure is fundamentally challenging in long-horizon trajectories as local evidence can be misleading. A tool call may be syntactically valid and return a plausible observation while encoding an assumption that becomes harmful only several turns later, whereas an incorrect late-stage action may simply be the consequence of an earlier wrong decision. Evaluating each intermediate step in isolation therefore gives an incomplete basis for supervision; reliable attribution requires reasoning over the full sequence of decisions, observations, and the final outcome.

We address this by generating feedback from the \emph{complete} failed rollout. Given a failed trajectory $\tau$, we instantiate the current policy $\pi_\theta$ as a hindsight analyzer $\mathcal{H}_\theta$, prompting it with the task, the full trajectory, and the instruction (as shown in \Cref{fig:prompt_multi_step}) to return a sparse set of failure-relevant steps together with corresponding corrective feedback:
\begin{equation}
\mathcal{H}_\theta(\tau) \mapsto \{(i, f_i)\}_{i \in \mathcal{I}},
\quad \mathcal{I} \subseteq \{1,\ldots,T\},
\end{equation}
where $\mathcal{I}$ denotes the set of selected failure-relevant steps, and $f_i$ is natural-language feedback describing why the action at step $i$ contributes to the failure and how it should have been corrected. Because the selection is made with global trajectory context, it can avoid supervising turns that are locally noisy or merely consequences of the root cause. The feedback generation stage therefore serves two roles at once: it produces corrective feedback and determines the target spans to which the correction should be applied.

\paragraph{Targeted Self-Distillation.}
With the selected failure-relevant steps $\mathcal{I}$, the remaining challenge is how to apply each correction to its corresponding action span without updating unrelated parts of the rollout. We resolve this by exploiting the information asymmetry inherent in \emph{self-distillation}. Rather than relying on an external supervisor or uniform reward signals, we leverage the policy itself as a localized expert by exposing it to hindsight feedback. For each identified step $i \in \mathcal{I}$, we augment the original interaction history $h_i$ with the generated feedback $f_i$, and query the teacher policy under this augmented context. Conditioning the policy on this privileged hindsight induces a locally improved teacher distribution, $\pi_\theta(\, \cdot \mid h_i, f_i, a_{i, <t})$, while the student remains conditioned only on the original history. We then minimize the reverse KL divergence between the two distributions only on the identified failure-relevant action spans:
\begin{equation}
\begin{aligned}
    \sum_{i\in\mathcal{I}} \sum_{t=1}^{\abs{a_i}} D_\mathrm{KL} \big(
    & \pi_\theta (\, \cdot \mid h_i,  a_{i, <t}) \parallel \\
    & \texttt{sg}(\pi_\theta (\, \cdot \mid h_i, f_i, a_{i, <t})) \big),
\end{aligned}
\label{eq:training_objective}
\end{equation}
where $\texttt{sg}(\cdot)$ denotes the stop-gradient. By deliberately narrowing the optimization landscape to these precise regions, the policy is encouraged to absorb high-quality feedback exactly where it erred. This targeted mechanism effectively provides dense, token-level supervision for long-horizon tasks with sparse rewards, enhancing learning efficiency while preserving original task performance by avoiding unnecessary updates to successful trajectories.

%% file: sections/3_experiment.tex
\section{Experiments}

\subsection{Experimental Setup}

\paragraph{Benchmarks \& Metrics.}
We evaluate \modelname{} on two complementary long-horizon agent benchmarks: BFCL v3~\citep{bfcl} and AppWorld~\citep{trivedi2024appworld}. BFCL v3 evaluates executable multi-turn function calling under schema and dialogue constraints; we use only the \textsc{Base} and \textsc{Long Context} categories from the multi-turn split. AppWorld evaluates stateful application workflows through Task Goal Completion, where agents interact with app APIs and are scored by unit tests over the final environment state. We run each task four times and report Avg@4 and Best@4.

\paragraph{Baselines.}
We compare \modelname{} against five baselines. \textbf{Initial} denotes the zero-shot policy before any intervention. \textbf{SFT} performs supervised fine-tuning on high-reward trajectories generated by GPT-5.4-mini~\citep{gpt54}. \textbf{GRPO}~\citep{grpo} optimizes terminal task rewards (without textual feedback) under the same rollout budget. \textbf{SDPO}~\citep{hubotter2026sdpo} conditions the teacher on hindsight feedback but distills the entire failed trajectory without target-turn selection. \textbf{OpenClaw-RL}~\citep{wang2026openclaw} uses next-state signals to derive scalar rewards and textual hints at each turn, testing dense local feedback without full-trajectory hindsight attribution. We report two variants of \modelname{}: \textbf{\modelname{}-Single}, which distills the first failure-relevant step, and \textbf{\modelname{}-Multi}, which distills multiple selected failure-relevant steps.

\paragraph{Implementation Details.}
We evaluate \modelname{} with Qwen3-4B-Instruct-2507~\citep{yang2025qwen3} as our primary backbone model. Across all rollout-based optimization methods, we use four rollouts per task and train for 15 epochs. For \modelname{}, we restrict hindsight feedback generation to at most three failure-relevant steps per failed trajectory. Additional details are provided in \Cref{sec:appendix_details}.

\subsection{Experimental Results}

\input{tables/main_result}
\input{figures/main_result}

\paragraph{Main Results.}
\Cref{tab:main_results} shows that \modelname{}-Multi achieves the best overall performance on both BFCL v3 and AppWorld. On BFCL v3, it improves Avg@4 from the strongest baseline score of 31.56 to 41.88 and Best@4 from 45.00 to 48.75. On AppWorld, it improves Avg@4 from 9.74 to 18.46 and Best@4 from 19.32 to 31.11. The baseline trends highlight the role of localization: GRPO improves on the initial policy but remains limited by sparse terminal rewards, full-trajectory SDPO benefits from hindsight feedback but can dilute corrective supervision across irrelevant or already-correct actions, and OpenClaw-RL achieves competitive Best@4 on BFCL v3 but lower Avg@4, suggesting that dense local hints are less stable across samples. In contrast, \modelname{} localizes feedback-conditioned distillation to failure-relevant turns. Even \modelname{}-Single, which distills only the first failure-relevant step, shows substantial gains over the baselines, demonstrating the efficacy of localized hindsight supervision. The additional gains from \modelname{}-Multi suggest that supervising multiple selected failure points extracts a richer corrective signal from the same rollout budget.

\paragraph{Training Dynamics and Efficiency.}
In addition to effectively providing dense supervision for relevant actions, \modelname{} is significantly more efficient than approaches that either supervise the entire trajectory or rely on per-step feedback or rewards. \Cref{fig:training_dynamics_efficiency}~(Left) shows that \modelname{} improves more rapidly and reaches the highest evaluation accuracy across training epochs, whereas GRPO and SDPO saturate at lower accuracies and OpenClaw-RL exhibits weaker stability. At the same time, because \modelname{} supervises only selected turns rather than every action or the full trajectory, it avoids much of the feedback-generation and distillation overhead of dense-feedback methods. \Cref{fig:training_dynamics_efficiency}~(Middle, Right) shows that this localization reduces time per training step from 84.76s to 37.45s and peak GPU memory from 126GB to 85GB, yielding a 2.26$\times$ lower step time and a 1.48$\times$ lower memory footprint than the dense per-turn feedback baseline. By limiting distillation to failure-relevant actions, \modelname{} improves optimization while substantially reducing training cost.

\input{tables/feedback_location}
\paragraph{Analysis of Target Turn Distribution.}
\Cref{fig:target_turn_distribution_over_training} shows that feedback targets are distributed across the trajectory rather than concentrated at the beginning. Across the first 15 epochs, 36.7\% of targets fall in turns 1--3, 44.8\% in turns 4--8, and 18.5\% in turn 9 or later. Notably, later targets (9+) increase from 14.0\% to 24.5\% over training, suggesting that feedback shifts toward later-stage corrections as early-stage errors are reduced. Since these corrections are often distant from the initial prompt, this motivates targeting feedback to the selected failure-relevant turn rather than treating it as a global trajectory-level hint.
  
\paragraph{Analysis of Feedback Placement.}
We examine whether applying hindsight feedback at the selected target turn improves rollout success. For each failed base-policy trajectory, the hindsight analyzer produces one feedback message and a target turn. We then run paired interventions with the same feedback: feedback is either inserted at the beginning of a fresh rollout or immediately before the target action after replaying the failed prefix. Each condition is compared against its no-feedback rollout, and feedback remains persistent in the context. \Cref{tab:feedback_location} shows that target-turn feedback yields larger success gains on both benchmarks, with Target -- Start gains of +5.99 points on BFCL v3 and +1.72 points on AppWorld. This suggests that the selected target turns are actionable and provide a stronger feedback-conditioned teacher signal than applying the same feedback globally from the start.

\input{tables/feedback_source}

\paragraph{Analysis of Feedback Source.}
To further analyze how the source of hindsight feedback affects \modelname{}, we compare different feedback sources in \Cref{tab:feedback_source}. Environmental feedback uses environment outputs directly rather than generating hindsight feedback, but it underperforms teacher-generated feedback variants. The EMA-updated teacher also consistently outperforms the fixed initial teacher, indicating that feedback generation benefits from tracking the improving policy. A larger teacher (GPT-5.4-mini) further improves performance, suggesting that stronger feedback can provide additional gains. Nevertheless, our EMA-updated teacher yields strong results without relying on an external large model, supporting the self-contained design of \modelname{}.

\paragraph{Analysis of Feedback Quality.}
We quantitatively assess whether the hindsight analyzer can localize both explicit and implicit failures. Across 631 failed BFCL v3 trajectories, \modelname{} identifies 765 target turns. GPT-5.4-mini~\citep{gpt54} classifies each selected turn by error type and, independently, identifies failure-relevant turns from the full trajectories to serve as a silver reference. As shown in \Cref{tab:feedback_quality}, 90.7\% of the selected targets are judged failure-relevant, comprising 58.2\% explicit execution failures and 32.5\% reasoning or state-tracking failures. Relative to the silver reference, localization agreement within $\pm1$ turn reaches 92.6\% for explicit failures and 85.1\% for implicit failures. These results show that the hindsight feedback generator captures not only readily observable tool-execution errors but also implicit reasoning and state-tracking failures, while accurately localizing the action spans most relevant for feedback-conditioned distillation.

\input{tables/feedback_quality}

\paragraph{Generalization across Backbones.}
We further evaluate \modelname{} on backbones of different scales, namely Qwen2.5-3B-Instruct and Qwen2.5-7B-Instruct~\citep{yang2024qwen25}. As shown in \Cref{tab:different_models}, \modelname{}-Multi consistently outperforms all baselines at both scales, improving BFCL v3 Avg@4 over the strongest baseline by 7.02 points on the 3B model and 12.18 points on the 7B model. The strong improvement on the smaller backbone demonstrates that the effectiveness of \modelname{} is not specific to the 4B model used in our main experiments. Moreover, the even larger gain on the 7B backbone suggests that \modelname{} can effectively leverage increased backbone capacity, highlighting its potential scalability across model sizes.

\input{tables/different_model}

%% file: tables/main_result.tex
\begin{table}[t]
\centering
\small
\setlength{\tabcolsep}{5pt}
\resizebox{\linewidth}{!}{
    \begin{tabular}{lcccc}
    \toprule
    & \multicolumn{2}{c}{\textbf{BFCL v3}} & \multicolumn{2}{c}{\textbf{AppWorld}}\\
    \cmidrule(lr){2-3} \cmidrule(lr){4-5}
    \textbf{Method} & Avg@4 & Best@4 & Avg@4 & Best@4 \\
    \midrule
    Initial & 25.94 & 36.25 & 5.98 & 13.85 \\
    SFT & 28.44 & 38.13 & 6.82 & 13.16 \\
    GRPO & 31.56 & 41.25 & 7.49 & 15.21 \\
    SDPO & 30.78 & 40.00 & 9.74 & 19.32 \\
    OpenClaw-RL & 28.28 & 45.00 & 7.65 & 12.31 \\
    \noalign{\vskip 0.25ex}\cdashline{1-5}\noalign{\vskip 0.75ex}
    \rowcolor{myblue!80}
    \textbf{\modelname{}-Single} & 36.25 & 43.13 & 16.54 & 29.40 \\
    \rowcolor{myblue!80}
    \textbf{\modelname{}-Multi} & \textbf{41.88} & \textbf{48.75} & \textbf{18.46} & \textbf{31.11} \\
    \bottomrule
    \end{tabular}
}
\vspace{-0.05in}
\caption{Comparison of \modelname{} and baselines on BFCL v3 and AppWorld. Best results are in \textbf{bold}.}
\vspace{-0.1in}
\label{tab:main_results}
\end{table}

%% file: figures/main_result.tex
\begin{figure*}[t]
\vspace{-0.15in}
    \centering
    \begin{minipage}[t]{0.69\textwidth}
        \centering
        \includegraphics[width=\linewidth]{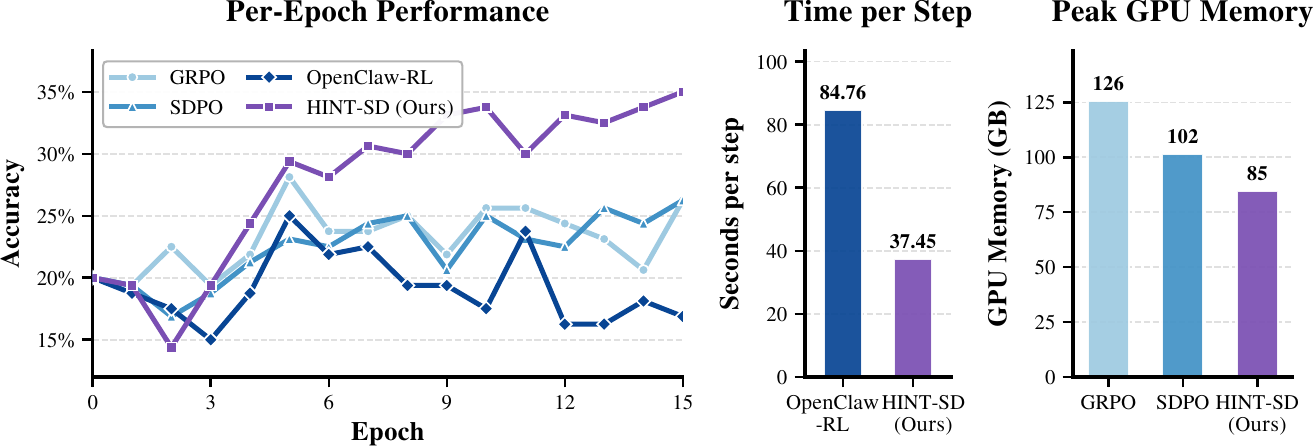}
        \caption{(Left) Per-epoch accuracy scores on the BFCL v3 eval split. (Middle) Time per training step. (Right) Peak GPU memory usage during the first epoch.}
        \label{fig:training_dynamics_efficiency}
    \end{minipage}
    \hfill
    \begin{minipage}[t]{0.29\textwidth}
        \centering
        \includegraphics[width=0.9\linewidth]{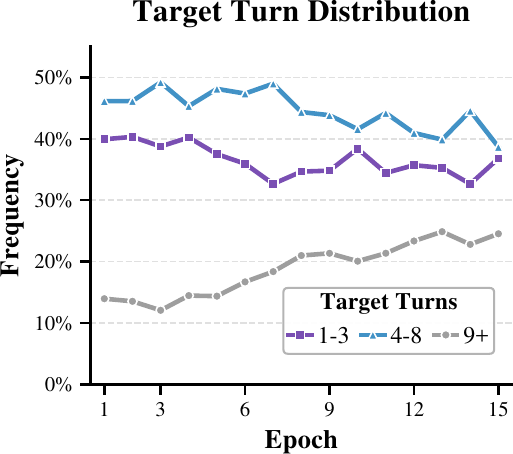}
        \caption{Distribution of selected target turns for BFCL v3.}
        \label{fig:target_turn_distribution_over_training}
    \end{minipage}
\vspace{-0.15in}
\end{figure*}

%% file: tables/feedback_location.tex
\begin{table}[t]
\centering
\small
\setlength{\tabcolsep}{4pt}
\resizebox{\linewidth}{!}{
    \begin{tabular}{lccc}
    \toprule
    \textbf{Benchmark} & \textbf{Start FB} & \textbf{Target FB} & \textbf{$\Delta$(Target--Start)} \\
    \midrule
    BFCL v3 & 2.68 & 8.67 & +5.99 \\
    AppWorld & 0.44 & 2.16 & +1.72 \\
    \bottomrule
    \end{tabular}
}
\vspace{-0.05in}
\caption{Comparison of feedback placements, showing percentage-point gains vs. the no-feedback rollout.}
\vspace{-0.2in}
\label{tab:feedback_location}
\end{table}

%% file: tables/feedback_source.tex
\begin{table}[t]
\centering
\small
\setlength{\tabcolsep}{5pt}
\resizebox{\linewidth}{!}{
    \begin{tabular}{lcccc}
    \toprule
    & \multicolumn{2}{c}{\textbf{BFCL v3}} & \multicolumn{2}{c}{\textbf{AppWorld}}\\
    \cmidrule(lr){2-3} \cmidrule(lr){4-5}
    \textbf{Feedback Source} & Avg@4 & Best@4 & Avg@4 & Best@4 \\
    \midrule
    \rowcolor{myblue!80}
    \textbf{Teacher (w/EMA)} & \underline{41.88} & \underline{48.75} & \underline{18.46} & \underline{31.11} \\
    Environment & 36.25 & 42.50 & 15.90 & 27.86 \\
    Initial Teacher & 37.50 & 45.63 & 14.40 & 28.89 \\
    Larger Teacher & \textbf{48.59} & \textbf{52.50} & \textbf{20.81} & \textbf{35.04} \\
    \bottomrule
    \end{tabular}
}
\vspace{-0.05in}
\caption{Comparison of different feedback sources.}
\vspace{-0.08in}
\label{tab:feedback_source}
\end{table}

%% file: tables/feedback_quality.tex
\begin{table}[t]
\centering
\small
\setlength{\tabcolsep}{4pt}
\resizebox{\linewidth}{!}{
    \begin{tabular}{lccc}
    \toprule
    \textbf{Error Type} & \textbf{Ratio} & \textbf{Exact Overlap} & \textbf{$\pm1$ Overlap} \\
    \midrule
    Explicit Execution & 58.2\% & 85.4\% & 92.6\% \\
    Reasoning/State & 32.5\% & 57.4\% & 85.1\% \\
    \bottomrule
    \end{tabular}
}
\vspace{-0.05in}
\caption{Error-type distribution of selected targets and silver-reference agreement on BFCL v3.}
\vspace{-0.1in}
\label{tab:feedback_quality}
\end{table}

%% file: tables/different_model.tex
\begin{table}[t]
\centering
\small
\setlength{\tabcolsep}{4pt}
\resizebox{\linewidth}{!}{
    \begin{tabular}{lcc}
    \toprule
    \textbf{Method} & \textbf{Qwen2.5-3B-Inst} & \textbf{Qwen2.5-7B-Inst} \\
    \midrule
    Initial & $8.59 \pm 1.02$ & $21.31 \pm 0.92$ \\
    SFT & $9.53 \pm 0.87$ & $24.79 \pm 0.82$ \\
    GRPO & $10.32 \pm 0.74$ & $26.05 \pm 0.85$ \\
    SDPO & $9.23 \pm 0.83$ & $25.44 \pm 1.34$ \\
    \noalign{\vskip 0.25ex}\cdashline{1-3}\noalign{\vskip 0.75ex}
    \rowcolor{myblue!80}
    \textbf{\modelname{}-Multi} & $\mathbf{17.34 \pm 0.76}$ & $\mathbf{38.23 \pm 0.59}$ \\
    \bottomrule
    \end{tabular}
}
\vspace{-0.05in}
\caption{BFCL v3 Avg@4 across backbone sizes, reported as mean and standard deviation.}
\vspace{-0.1in}
\label{tab:different_models}
\end{table}

%% file: sections/4_conclusion.tex
\section{Conclusion}

We presented \modelname{}, a targeted hindsight self-distillation framework for long-horizon LLM agents. Rather than applying dense feedback uniformly across an entire failed trajectory, \modelname{} uses full-trajectory hindsight to identify failure-relevant turns and distills a feedback-conditioned teacher only at those selected action spans. Experiments on BFCL v3 and AppWorld show that this targeted formulation improves performance and training efficiency over reward-only optimization, full-trajectory distillation, and dense turn-level feedback baselines. Our target-turn and feedback-placement analyses further indicate that selected hindsight targets are distributed across trajectories and provide more actionable supervision when applied at the relevant turn. Overall, these results suggest that deciding where to apply feedback is a central design choice for effective and efficient long-horizon agent post-training.

%% file: sections/X_limitations.tex
\section*{Limitations}

While \modelname{} effectively generates hindsight feedback for targeted self-distillation, its training signal still depends on whether the generated feedback correctly identifies actionable failures and proposes corrections that improve task completion. This requires the initial model to have sufficient instruction-following and task-solving capability to reason about failed trajectories. Nevertheless, our results suggest that effective feedback generation remains possible even at smaller model scales. Future work can further guide feedback generation with additional supervision or constraints to improve feedback quality.

%% file: sections/X_ethical_considerations.tex
\section*{Ethical Considerations}

\modelname{} leverages the model itself as both a hindsight feedback generator and a feedback-conditioned teacher, enabling the agent to improve through self-generated feedback without relying on stronger external supervision. By focusing hindsight feedback on failure-relevant action spans rather than every step in a trajectory, the approach can make long-horizon agent training more efficient and accessible. However, when connected to real-world tools, the agent may execute incorrect, unauthorized, or harmful actions. Because \modelname{} relies on the model's own feedback signals, such harmful behaviors may also reinforce existing biases or failure modes, highlighting the need for appropriate safeguards before real-world deployment.

%% file: sections/X_acknowledgements.tex
\section*{Acknowledgements}

This work was supported by the Institute of Information \& communications Technology Planning \& Evaluation (IITP) grant funded by the Korea government (MSIT) (RS-2019-II190075, Artificial Intelligence Graduate School Program (KAIST)), the National Research Foundation of Korea (NRF) grant funded by the Korea government (MSIT) (RS-2023-00256259), the IITP grant funded by the Ministry of Science and ICT (MSIT) of the Republic of Korea in connection with the Global AI Frontier Lab International Collaborative Research (RS-2024-00469482 \& RS-2024-00509279), the IITP grant funded by the Korea government (MSIT) (RS-2022-II220713, Meta-learning Applicable to Real-world Problems), and the Advanced GPU Utilization Support Program funded by the Government of the Republic of Korea (MSIT).

%% file: sections/X_appendix.tex
\clearpage
\appendix

\section{Additional Experimental Details}
\label{sec:appendix_details}

\subsection{Additional Details on Baselines}

\paragraph{SFT.} The SFT baseline is trained on demonstrations generated by GPT-5.4-mini. For each training task, we sample up to 10 candidate trajectories from the teacher model and execute them in the corresponding environment. If at least one candidate satisfies the benchmark-specific success criterion, we retain a successful trajectory as the demonstration. Otherwise, we select the trajectory with the highest reward, providing the strongest available supervision for that task. We then fine-tune the student model using step-wise teacher forcing. At each interaction step $t$, the student conditions on the trajectory prefix $h_t$ and is trained to predict the corresponding assistant action $a_t$.

\paragraph{GRPO.} The GRPO~\citep{grpo} baseline follows the standard GRPO objective, where rewards are computed solely from the final environment state. Because intermediate interaction steps cannot be directly evaluated or assigned step-level rewards, we optimize a trajectory-level objective using the global terminal reward. Concretely, the same scalar reward is propagated uniformly across all action spans $a_1, \dots, a_T$ within the trajectory.

\paragraph{SDPO.} Although SDPO~\citep{hubotter2026sdpo} was originally proposed for single-turn settings, we extend it to the multi-turn setting. After each rollout, we use a teacher model to generate natural-language feedback conditioned on the final trajectory and outcome. This global feedback is then prepended to the initial prompt as privileged context, and the feedback-conditioned teacher distills all assistant action spans in the failed trajectory without target-turn selection.

\paragraph{OpenClaw-RL.} OpenClaw-RL~\citep{wang2026openclaw} trains agents from next-state signals observed immediately after each action. The judge converts each next-state signal into a turn-level scalar reward and textual feedback. The reward is used for policy optimization, while the feedback is inserted only into the teacher context for feedback-conditioned distillation. Unlike \modelname{}, however, OpenClaw-RL provides dense local supervision rather than using full-trajectory hindsight. Its immediate next-state judge can therefore be misaligned with failures that emerge several turns after an earlier planning or state-tracking error.

\subsection{Additional Implementation Details}

For BFCL v3, we split the full task set into train/eval/test partitions with a ratio of 5:1:4. For optimization, we use AdamW~\citep{adamw} with a base learning rate of $5\times10^{-6}$ for BFCL v3 and $3\times10^{-6}$ for AppWorld, together with a linear scheduler and warm-up over the first $5\%$ of training steps. We apply LoRA~\citep{lora} to the query and value projection layers with rank $r=32$, scaling factor $\alpha=64$, and dropout rate~\citep{dropout} of $0.05$. Teacher parameters are initialized from the student and updated via EMA~\citep{ema} with an update rate of $0.001$. We select the checkpoint with the highest reward on the evaluation split. We implement all optimization-based methods with TRL~\citep{trl} and use vLLM~\citep{vllm} for efficient on-policy generation. All experiments are conducted on a single NVIDIA H200 GPU. The feedback-generation prompts for the \modelname{}-Multi and \modelname{}-Single variants are shown in \Cref{fig:prompt_multi_step,fig:prompt_single_step}.

\section{Additional Experimental Results}
\subsection{Distribution of Selected Hindsight Targets}
\label{sec:appendix_target_selection}

\Cref{fig:target_turn_distribution} aggregates selected hindsight target turns, complementing the epoch-wise regions in \Cref{fig:target_turn_distribution_over_training}. Targets concentrate in early and middle interactions (mean turn 5.32), but are not restricted to the start: 10.0\% occur after turn 10. This supports the relevance-sparsity view behind \modelname{}: hindsight supervision should be attached to selected failure-relevant turns rather than uniformly to the full trajectory or always from the beginning.

\input{figures/target_turn_distribution}

\subsection{Stability across Repeated Runs}

\input{tables/main_result_stdev}

We additionally report standard deviations across repeated runs in \Cref{tab:main_results_stdev}, extending the results in \Cref{tab:main_results}. Across both benchmarks, standard deviations are at most 0.92, and \modelname{}-Multi maintains substantial margins over the strongest baselines. These results confirm that the gains of \modelname{} are consistent across runs.

\subsection{Example of Targeted Hindsight Feedback}
\label{sec:appendix_qualitative_target}

\Cref{fig:qualitative_target_feedback} illustrates how target-turn selection makes hindsight feedback action-specific: rather than giving a generic episode-level summary, the analyzer specifies how each selected action should change. For example, it identifies the missing Spotify access token at Turn 14 and the undefined token variable at Turn 15, making the feedback directly actionable at the selected action spans. \Cref{fig:qualitative_feedback_comparison_bfcl,fig:qualitative_feedback_comparison_appworld} further compare global hindsight feedback with \modelname{} multi-step feedback, showing that global feedback summarizes the episode-level failure while selected-turn feedback attaches corrections to the concrete actions where the failure becomes actionable.

\input{sections/X_related_works}

\clearpage
\input{figures/prompt}

\clearpage
\input{figures/qualitative_hint_multi}

\clearpage
\input{figures/qualitative_feedback_comparison}

%% file: figures/target_turn_distribution.tex
\begin{figure}[b]
    \centering
    \includegraphics[width=0.7\linewidth]{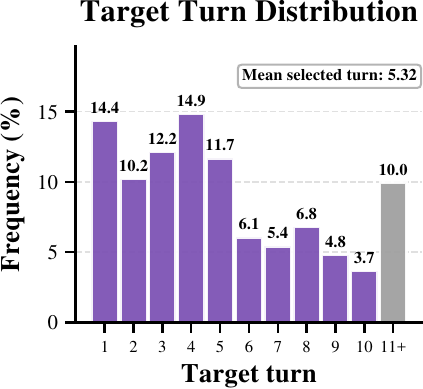}
    \caption{Count distribution of selected hindsight target turns on BFCL v3 over the first 15 training epochs. Turns after 10 are aggregated into the 11+ bin.}
    \label{fig:target_turn_distribution}
\end{figure}

%% file: tables/main_result_stdev.tex
\begin{table}[h]
\centering
\small
\setlength{\tabcolsep}{6pt}
\resizebox{\linewidth}{!}{
    \begin{tabular}{lcc}
    \toprule
    \textbf{Method} & \textbf{BFCL v3} & \textbf{AppWorld} \\
    \midrule
    Initial & $24.84 \pm 0.68$ & $5.77 \pm 0.76$ \\
    SFT & $28.13 \pm 0.72$ & $6.21 \pm 0.56$ \\
    GRPO & $30.97 \pm 0.63$ & $7.59 \pm 0.81$ \\
    SDPO & $30.23 \pm 0.85$ & $8.32 \pm 0.92$ \\
    OpenClaw-RL & $29.13 \pm 0.90$ & $7.83 \pm 0.67$ \\
    \noalign{\vskip 0.25ex}\cdashline{1-3}\noalign{\vskip 0.75ex}
    \rowcolor{myblue!80}
    \textbf{\modelname{}-Single} & $35.85 \pm 0.82$ & $17.02 \pm 0.83$ \\
    \rowcolor{myblue!80}
    \textbf{\modelname{}-Multi} & $\mathbf{42.32 \pm 0.71}$ & $\mathbf{18.39 \pm 0.79}$ \\
    \bottomrule
    \end{tabular}
}
\vspace{-0.05in}
\caption{Avg@4 results on BFCL v3 and AppWorld, reported as mean and standard deviation.}
\vspace{-0.1in}
\label{tab:main_results_stdev}
\end{table}

%% file: sections/X_related_works.tex
\section{Related Work}

\paragraph{Credit assignment and selective training.}
Long-horizon agent training requires assigning sparse outcome signals to intermediate decisions. Work on verifiers and process supervision~\citep{cobbe2021verifiers, lightman2023verify} shows that intermediate labels can be more informative than final outcomes alone. AgentEvolver~\citep{agentevolver} uses LLM-based self-attribution to score each action's contribution to the final outcome and converts these scores into process-level rewards for GRPO optimization. Other long-horizon methods select or reweight informative states. PivotRL~\citep{yi2026pivotrl} trains on informative pivots from expert trajectories, GiGPO~\citep{xie2025gigpo} targets fine-grained credit assignment in group-based RL, and HCAPO~\citep{tan2026hcapo} uses hindsight reasoning to refine step-level credit for policy optimization. These methods can identify failure-causing actions, but the resulting signal is typically scalar or policy-gradient-based, so learning the correct alternative action still depends on sparse successful rollouts.

\paragraph{Feedback-Conditioned Distillation.}
Natural-language and environment feedback has been used to revise model outputs and provide privileged training signals. Reflexion~\citep{shinn2023reflexion}, Self-Refine~\citep{madaan2023selfrefine}, and CRITIC~\citep{gou2024critic} use verbal or tool-grounded feedback for iterative correction. SDPO~\citep{hubotter2026sdpo} and RLTF~\citep{song2026rltf} instead internalize such feedback by distilling feedback-conditioned behavior into a feedback-free policy. In agent settings, OpenClaw-RL~\citep{wang2026openclaw} converts next-state signals into turn-level rewards or textual hints, and Skill-SD~\citep{wang2026skillsd} conditions the teacher on retrieved skill descriptions. However, these methods either treat feedback as trajectory-level supervision or analyze feedback at every turn, leaving open which agent action should receive the corrective signal and introducing unnecessary cost when most turns are already correct or irrelevant. \modelname{} instead uses full-trajectory hindsight to select failure-relevant action spans and applies feedback-conditioned distillation only at those targeted turns.

%% file: figures/prompt.tex
\begin{figure*}[t]
    \centering
\begin{minipage}{0.8\textwidth}
\begin{promptbox}[]
SYSTEM:
You analyze failed AppWorld tool-use trajectories. Identify up to {max_steps} problematic steps where the agent made mistakes (between 1 and {max_steps} steps total, ordered earliest first). For EACH problematic step, write a distinct correction in less than three sentences that targets only that step's mistake. Output valid JSON only, no other text.

USER:
Review this failed AppWorld attempt and identify the problematic steps. Return one (step, feedback) pair per problematic step.

{trajectory}

Output format (JSON only, no other text):
{"failures": [{"step": <1-indexed step number>, "feedback": "<correction for this step>"}, ...]}
\end{promptbox}
    \caption{Prompt template for multi-step hindsight feedback generation in \modelname{}-Multi. Given a complete failed AppWorld trajectory, the analyzer selects up to \texttt{\{max\_steps\}} failure-relevant steps and returns localized corrective feedback for each selected step.}
    \label{fig:prompt_multi_step}
\end{minipage}
\end{figure*}

\begin{figure*}[t]
    \centering
\begin{minipage}{0.8\textwidth}
\begin{promptbox}[]
SYSTEM:
You analyze failed AppWorld tool-use trajectories. Identify the FIRST step where the agent made a mistake. Write the feedback in less than three sentences. Output valid JSON only, no other text.

USER:
Review this failed AppWorld attempt and identify the first problematic step.

{trajectory}

Output format (JSON only, no other text):
{"failure_step": <1-indexed step number>, "feedback": "<correction sentence(s)>"}
\end{promptbox}
    \caption{Prompt template for single-step hindsight feedback generation in \modelname{}-Single. Given a complete failed AppWorld trajectory, the analyzer identifies the earliest failure-relevant step and returns a concise correction for the step.}
    \label{fig:prompt_single_step}
\end{minipage}
\end{figure*}

%% file: figures/qualitative_hint_multi.tex
\begin{figure*}[t]
\centering
\begin{minipage}{\textwidth}

\begin{tcolorbox}[targettaskstyle,title={Task}]
Start playing a playlist on Spotify that has enough songs for my workout today.
I do not want to have to change the playlist in the middle of my workout.
The workout plan is in Simple Note.
\end{tcolorbox}\vspace{-4pt}

\begin{tcolorbox}[targettrajstyle,title={Abbreviated Execution Trajectory}]
\textcolor{targettoolcall}{\textbf{[Turns 1--13 (Summarized)]}}\\
The agent retrieves account information, logs into Simple Note, searches for the workout note, and parses the workout duration for the current day. It then moves to Spotify and obtains a login response stored as \texttt{login\_result}.\\[5pt]
\textcolor{targettoolcall}{\textbf{[Selected Turn 14 Action]}}\\
playlists = apis.spotify.search\_playlists(query="workout")\\[3pt]
\textcolor{targettoolresult}{\textbf{[Observation]}}\\
Exception: Response status code is 401:\\
\{"message": "You are either not authorized to access this spotify API endpoint\\
\hspace{1em}or your access token is missing, invalid or expired."\}\\[6pt]
\textcolor{targettoolcall}{\textbf{[Selected Turn 15 Action]}}\\
playlists = apis.spotify.search\_playlists(\\
\hspace{1em}query="workout",\\
\hspace{1em}access\_token=spotify\_access\_token["access\_token"]\\
)\\[3pt]
\textcolor{targettoolresult}{\textbf{[Observation]}}\\
NameError: name 'spotify\_access\_token' is not defined\\[5pt]
\textcolor{targettoolcall}{\textbf{[Turns 16--20 (Summarized)]}}\\
The rollout continues after the invalid Spotify state, but no valid playlist playback is completed before termination.
\end{tcolorbox}\vspace{-4pt}

\begin{tcolorbox}[targetfeedbackstyle,title={Selected Hindsight Feedback}]
\textbf{Turn 14.} The API call is missing the access token; add it as a parameter in the \texttt{search\_playlists} call.

\textbf{Turn 15.} The variable \texttt{spotify\_access\_token} is undefined; use \texttt{login\_result} from the previous step instead.
\end{tcolorbox}
\end{minipage}
\caption{Qualitative example of selected hindsight target turns from an AppWorld training rollout. The abbreviated trajectory context shows that the analyzer localizes feedback to the actions where the agent loses the authenticated Spotify state, rather than applying the same feedback globally at the beginning of the trajectory.}
\label{fig:qualitative_target_feedback}
\end{figure*}

%% file: figures/qualitative_feedback_comparison.tex
\begin{figure*}[t]
\centering
\begin{minipage}{\textwidth}

\begin{tcolorbox}[targettaskstyle,title={Task}]
\textbf{BFCL travel booking.}
Book an economy flight from Crescent Hollow to New York on November 12, 2026 using the user's card and access token.
The conversation then asks for travel insurance, an invoice, and customer-support handling for the same booking.
\end{tcolorbox}

\begin{tcolorbox}[targettrajstyle,title={Abbreviated Execution Trajectory}]
\textcolor{targettoolcall}{\textbf{[Turns 1--2 (Summarized)]}}\\
The agent maps Crescent Hollow to CRH and New York to JFK.\\[5pt]
\textcolor{targettoolcall}{\textbf{[Turn 3 Action]}}\\
get\_flight\_cost(\\
\hspace{1em}travel\_from="CRH",\\
\hspace{1em}travel\_to="JFK",\\
\hspace{1em}travel\_date="2026-11-12",\\
\hspace{1em}travel\_class="economy"\\
)\\[3pt]
\textcolor{targettoolresult}{\textbf{[Observation]}}\\
\{"travel\_cost\_list": [300.0]\}\\[6pt]
\textcolor{targettoolcall}{\textbf{[Selected Turn 4 Action]}}\\
The assistant explains the mismatch between the expected \$850 fee and the observed \$300 cost, asks for confirmation, and does not call book\_flight.\\[3pt]
\textcolor{targettoolresult}{\textbf{[Next User]}}\\
The user refers to the trip as post-booking and asks to purchase \$100 travel insurance.\\[6pt]
\textcolor{targettoolcall}{\textbf{[Selected Turn 5 Action]}}\\
purchase\_insurance(\\
\hspace{1em}access\_token=...,\\
\hspace{1em}booking\_id="bk\_12345",\\
\hspace{1em}card\_id="card\_6789",\\
\hspace{1em}insurance\_cost=100\\
)\\[3pt]
\textcolor{targettoolresult}{\textbf{[Observation]}}\\
\{"insurance\_status": false, "error": "Booking not found"\}\\[6pt]
\textcolor{targettoolcall}{\textbf{[Selected Turn 7 Action]}}\\
retrieve\_invoice(\\
\hspace{1em}access\_token=...,\\
\hspace{1em}booking\_id="bk\_12345",\\
\hspace{1em}insurance\_id="ins\_98765"\\
)\\[3pt]
\textcolor{targettoolresult}{\textbf{[Observation]}}\\
\{"error": "Booking not found"\}
\end{tcolorbox}\vspace{-4pt}

\begin{tcolorbox}[targetfeedbackstyle,title={Feedback Comparison}]
\textbf{Global hindsight.} The episode-level feedback identifies one root cause: the assistant never created a valid booking, so the later insurance, invoice, and support calls all used a non-existent booking id. It recommends calling \texttt{book\_flight} before any booking-dependent action.

\textbf{\modelname{} multi-step.} \textbf{Turn 4:} initiate the booking with \texttt{book\_flight}. \textbf{Turn 5:} do not buy insurance with the fabricated booking id \texttt{bk\_12345}. \textbf{Turn 7:} retrieve the invoice only after a successful booking creates a valid booking id.
\end{tcolorbox}
\end{minipage}
\caption{Qualitative comparison on a BFCL task. The task-matched rollouts share the same failure pattern: the booking is never created, so later booking-dependent tool calls fail. Global hindsight gives one episode-level correction, while \modelname{} attaches the same root cause to the concrete turns where it first appears and then propagates.}
\label{fig:qualitative_feedback_comparison_bfcl}
\end{figure*}

\begin{figure*}[t]
\centering
\begin{minipage}{\textwidth}

\begin{tcolorbox}[targettaskstyle,title={Task}]
\textbf{AppWorld file system.}
Compress each vacation-spot subdirectory under \texttt{\textasciitilde/photos/} into a tar file under \texttt{\textasciitilde/photos/vacations/}, then delete the original vacation-spot subdirectories.
\end{tcolorbox}\vspace{-4pt}

\begin{tcolorbox}[targettrajstyle,title={Abbreviated Execution Trajectory}]
\textcolor{targettoolcall}{\textbf{[Selected Turn 1 Action]}}\\
import os\\
import tarfile\\
Use local filesystem calls to list vacation folders under \textasciitilde/photos/.\\[3pt]
\textcolor{targettoolresult}{\textbf{[Observation]}}\\
Usage of the following module is not allowed: tarfile.\\[6pt]
\textcolor{targettoolcall}{\textbf{[Turns 2--4 (Summarized)]}}\\
The agent inspects available apps and file\_system API descriptions, which list show\_directory rather than list\_directory.\\[5pt]
\textcolor{targettoolcall}{\textbf{[Selected Turn 3 Action]}}\\
vacation\_spots = apis.file\_system.list\_directory("\textasciitilde/photos/")\\[3pt]
\textcolor{targettoolresult}{\textbf{[Observation]}}\\
Exception: No API named list\_directory found in the file\_system app.\\[6pt]
\textcolor{targettoolcall}{\textbf{[Selected Turn 5 Action]}}\\
vacation\_spots = apis.file\_system.show\_directory("\textasciitilde/photos/")\\[3pt]
\textcolor{targettoolresult}{\textbf{[Observation]}}\\
TypeError: positional argument error. The API specification later shows that show\_directory expects named arguments, including access\_token.
\end{tcolorbox}\vspace{-4pt}

\begin{tcolorbox}[targetfeedbackstyle,title={Feedback Comparison}]
\textbf{Global hindsight.} The task-matched episode feedback focuses on a later login failure and recommends using the full email address when calling \texttt{apis.file\_system.login}.

\textbf{\modelname{} multi-step.} \textbf{Turn 1:} avoid disallowed system modules and use environment APIs. \textbf{Turn 3:} replace \texttt{list\_directory} with the documented \texttt{show\_directory} API. \textbf{Turn 5:} authenticate first and pass \texttt{access\_token} as a named parameter.
\end{tcolorbox}
\end{minipage}
\caption{Qualitative comparison on an AppWorld task. The trajectory excerpt is from the selected-target run. The selected-turn feedback exposes early actionable errors in API use and authentication, instead of only summarizing a later episode-level failure.}
\label{fig:qualitative_feedback_comparison_appworld}
\end{figure*}